\pdfoutput=1

\documentclass[final,3p,authoryear,times,twocolumn]{elsarticle}

\usepackage{graphics}
\usepackage{amssymb}
\usepackage{amsmath}
\usepackage[table,xcdraw]{xcolor}
\usepackage{booktabs}
\usepackage{subcaption}

\usepackage{array}
\newcolumntype{L}[1]{>{\raggedright\let\newline\\\arraybackslash\hspace{0pt}}m{#1}}
\newcolumntype{C}[1]{>{\centering\let\newline\\\arraybackslash\hspace{0pt}}m{#1}}
\newcolumntype{R}[1]{>{\raggedleft\let\newline\\\arraybackslash\hspace{0pt}}m{#1}}

\usepackage[colorlinks=true,linkcolor=blue,citecolor=blue,urlcolor=blue,pagebackref=false]{hyperref}
\usepackage{url}
\usepackage{epstopdf}
\journal{Preprint submitted to Elsevier}

\begin{document}

\begin{frontmatter}

\title{Validation, comparison, and combination of algorithms for automatic detection of pulmonary nodules in computed tomography images: The LUNA16 challenge}

\author[DIAG]{Arnaud~Arindra~Adiyoso~Setio\fnref{equal}}
\author[Turin,INFN_Turin,DIAG]{Alberto Traverso\fnref{equal}}
\author[RU_ICIS]{Thomas~de~Bel}
\author[RU]{Moira~S.N.~Berens}
\author[RU_ICIS]{Cas~van~den~Bogaard}
\author[INFN_Turin]{Piergiorgio~Cerello}
\author[CUMEDVIS]{Hao~Chen}
\author[CUMEDVIS]{Qi~Dou}
\author[Pisa_physics,INFN_Pisa]{Maria~Evelina~Fantacci}
\author[RAD]{Bram~Geurts} 
\author[RU]{Robbert~van~der~Gugten}
\author[CUMEDVIS]{Pheng~Ann~Heng}
\author[ETRO,IMEC]{Bart~Jansen}
\author[RU]{Michael~M.J.~de~Kaste}
\author[RU_ICIS]{Valentin~Kotov}
\author[jack_lin]{Jack~Yu-Hung~Lin}
\author[RU]{Jeroen~T.M.C.~Manders}
\author[CBM,ETRO,IMINDS]{Alexander~S\'o\~nora-Mengana}
\author[CBM]{Juan~Carlos~Garc\'ia-Naranjo}
\author[ETRO]{Evgenia~Papavasileiou}
\author[DIAG]{Mathias~Prokop}
\author[INFN_Turin]{Marco~Saletta}
\author[DIAG,AMES]{Cornelia~M~Schaefer-Prokop}
\author[DIAG]{Ernst~T.~Scholten}
\author[RU_ICIS]{Luuk~Scholten}
\author[RAD]{Miranda~M.~Snoeren}
\author[CEADEN]{Ernesto~Lopez~Torres}
\author[ETRO,IMEC]{Jef~Vandemeulebroucke}
\author[RU_ICIS]{Nicole~Walasek}
\author[RU]{Guido~C.A.~Zuidhof}
\author[DIAG,MEVIS]{Bram~van~Ginneken}
\author[DIAG]{Colin~Jacobs}

\address[DIAG]{Diagnostic Image Analysis Group, Radboud University Medical Center, Nijmegen, The Netherlands}
\address[RAD]{Department of Radiology and Nuclear Medicine, Radboud University Medical Center, Nijmegen, The Netherlands}
\address[AMES]{Department of Radiology, Meander Medisch Centrum, Amersfoort, The Netherlands}
\address[CBM]{Centro de Biof\'isica M\'edica, Universidad de Oriente, Santiago de Cuba, Cuba}
\address[ETRO]{Department of Electronics and Informatics, Vrije Universiteit Brussel, Brussels, Belgium}
\address[IMEC]{imec, Leuven, Belgium}
\address[Turin]{Department of Applied Science and Technology, Polytechnic University of Turin, Turin, Italy}
\address[INFN_Turin]{Turin Section of Istituto Nazionale di Fisica Nucleare, Turin, Italy}
\address[INFN_Pisa]{Pisa Section of Istituto Nazionale di Fisica Nucleare, Pisa, Italy}
\address[jack_lin]{Yan'an Xi Lu 129, 9th floor, Shanghai, China}
\address[Pisa_physics]{Department of Physics, University of Pisa, Pisa, Italy}
\address[CEADEN]{Center of Applied Technologies and Nuclear Development, La Habana, Cuba}
\address[CUMEDVIS]{Department of Computer Science and Engineering, The Chinese University of Hong Kong, China}
\address[RU]{Radboud University, Nijmegen, The Netherlands}
\address[RU_ICIS]{Institute for Computing and Information Sciences, Radboud University Nijmegen, Nijmegen, The Netherlands}
\address[MEVIS]{Fraunhofer MEVIS, Bremen, Germany}

\fntext[equal]{These authors contributed equally to this work}

\begin{abstract}
Automatic detection of pulmonary nodules in thoracic computed tomography (CT) scans has been an active area of research for the last two decades. 
However, there have only been few studies that provide a comparative performance evaluation of different systems on a common database. We have therefore set up the LUNA16 challenge, an objective evaluation framework for automatic nodule detection algorithms using the largest publicly available reference database of chest CT scans, the LIDC-IDRI data set.
In LUNA16, participants develop their algorithm and upload their predictions on 888 CT scans in one of the two tracks: 1) the complete nodule detection track where a complete CAD system should be developed, or 2) the false positive reduction track where a provided set of nodule candidates should be classified.
This paper describes the setup of LUNA16 and presents the results of the challenge so far. Moreover, the impact of combining individual systems on the detection performance was also investigated. It was observed that the leading solutions employed convolutional networks and used the provided set of nodule candidates. The combination of these solutions achieved an excellent sensitivity of over 95\% at fewer than 1.0 false positives per scan. This highlights the potential of combining algorithms to improve the detection performance. Our observer study with four expert readers has shown that the best system detects nodules that were missed by expert readers who originally annotated the LIDC-IDRI data. We released this set of additional nodules for further development of CAD systems.
\end{abstract}


\end{frontmatter}

\newpage
\section{Introduction}
\label{sec:intro}
Lung cancer is the deadliest cancer worldwide, accounting for approximately 27\% of cancer-related deaths in the United States (\cite{ACRStat16}). The NLST trial showed that three annual screening rounds of high-risk subjects using low-dose computed tomography (CT) reduced lung cancer mortality after 7 years by 20\% in comparison to screening with chest radiography (\cite{Aber11}). As a result of this trial and subsequent modeling studies, lung cancer screening programs using low-dose CT are currently being implemented in the U.S. and other countries will likely follow soon. One of the major challenges arising from the implementation of these screening programs is the enormous amount of CT images that must be analyzed by radiologists.

In the last two decades, researchers have been developing Computer-Aided-Detection (CAD) systems for automatic detection of pulmonary nodules. CAD systems are intended to make the interpretation of CT images faster and more accurate, hereby improving the cost-effectiveness of the screening program. The typical setup of a CAD system consists of: 1) preprocessing, 2) nodule candidate detection, and 3) false positive reduction. Preprocessing is typically used to standardize the data, restrict the search space for nodules to the lungs, and reduce noise and image artifacts. The candidate detection stage aims to detect nodule candidates at a very high sensitivity, which typically comes with many false positives. Subsequently, the false positive reduction stage reduces the number of false positives among the candidates and generates the final set of CAD marks.

Although a large number of CAD systems have been proposed (\cite{Berg16a,Torr15,Ginn15,Brow14,Jaco14,Choi13,Tan13b,Tera13,Casc12,Guo12,Cama11,Tan11e,Ricc11,Mess10a,Golo09,Murp09}), there have only been few studies providing an objective comparative evaluation framework using a common database.
The reported performances of published CAD systems can vary substantially because different data sets were used for training and evaluation (\cite{Firm14, Jaco15a}). Moreover, substantial variability among radiologists on what constitutes a nodule has been reported (\cite{Arma09}). Consequently, it is difficult to directly and objectively compare different CAD systems. The evaluation of different systems using the same framework provides unique information that can be leveraged to further improve the existing systems and develop novel solutions. 

ANODE09 was the first comparative study aimed towards evaluating nodule detection algorithms (\cite{Ginn10a}). This challenge has allowed groups to evaluate their algorithms on a shared set of scans obtained from a lung cancer screening trial. However, this study only included 50 scans from a single center, all acquired using one type of scanner and scan protocol. In addition, the ANODE09 set contained a limited number of larger nodules, which generally have a higher suspicion of malignancy. Evaluation on a larger and more diverse image database is therefore needed.

In this paper, we introduce a novel evaluation framework for automatic detection of nodules in CT images. A large data set, containing 888 CT scans with annotations from the publicly available LIDC-IDRI database (\cite{Arma11}), is provided for both training and testing. A web framework has been developed to efficiently evaluate algorithms and compare the result with the other algorithms. The impact of combining multiple candidate detection approaches and false positive reduction stages was also evaluated.

The key contributions of this paper are as follows. (1) We describe and provide an objective web framework for evaluating nodule detection algorithms using the largest publicly available data set; (2) We report the performance of algorithms submitted to the framework and investigate the impact of combining individual algorithms on the detection performance. We show that the combination of classical candidate detectors and a combination of deep learning architectures processing these candidates generates excellent results, better than any individual system; (3) We update the LIDC-IDRI reference standard by identifying nodules that were missed in the original LIDC-IDRI annotation process.

\section{Data}
\label{sec:data}

The data set was collected from the largest publicly available reference database for lung nodules: the LIDC-IDRI (\cite{Arma11,Clar13,Arma15}). This database is available from NCI's Cancer Imaging Archive\footnote{\url{https://wiki.cancerimagingarchive.net/display/Public/LIDC-IDRI}} under a Creative Commons Attribution 3.0 Unsupported License. The LIDC-IDRI database contains a total of 1018 CT scans. CT images come with associated XML files with annotations from four experienced radiologists. The database is very heterogeneous: It consists of clinical dose and low-dose CT scans collected from seven different participating academic institutions, and a wide range of scanner models and acquisition parameters.

As recommended by \cite{Naid13,Mano14} and the American College of Radiology (\cite{Kaze14}), thin-slice CT scans should be used for the management of pulmonary nodules. Therefore, we discarded scans with a slice thickness greater than 3~mm. On top of that, scans with inconsistent slice spacing or missing slices were also excluded. This led to the final list of 888~scans. These scans were provided as MetaImage (.mhd) images that can be accessed and downloaded from the LUNA16 website\footnote{\url{https://luna16.grand-challenge.org/}}. A more extensive data set description is provided in our previous study~(\cite{Jaco15a}).

Each LIDC-IDRI scan was annotated by experienced thoracic radiologists in a two-phase reading process. In the initial blinded reading phase, four radiologists independently annotated scans and marked all suspicious lesions as: nodule $\geq$~3~mm; nodule $<$~3~mm; non-nodule (any other pulmonary abnormality). For lesions annotated as nodule $\geq$~3~mm, diameter measurements were provided. In a subsequent unblinded reading phase, the anonymized blinded results of all other radiologists were revealed to each radiologist, who then independently reviewed all marks. No consensus was forced.

In the 888 scans, a total of 36,378 annotations were made by the radiologists. We only considered annotations categorized as nodules $\geq$~3~mm as relevant lesions, as nodules $<$~3~mm, and non-nodule lesions are not considered relevant for lung cancer screening protocols (\cite{Aber11}). Nodules could be annotated by multiple radiologists; annotations from different readers that were located closer than the sum of their radii were merged. In this case, position and diameters of these merged annotations were averaged. 
This resulted in a set of 2,290, 1,602, 1,186, and 777 nodules annotated by at least 1, 2, 3, or 4 radiologists, respectively.
We considered the 1,186 nodules annotated by the majority of the radiologists (at least 3 out of 4 radiologists) as the positive examples in our reference standard. These are the lesions that the algorithms should detect.
Other findings (1,104~nodules annotated by less than 3 out of 4 radiologists, 11,509~``nodule $<$~3~mm'' annotations, and 19,004~``non-nodule'' annotations) were considered ``irrelevant findings'' and marks on these locations were not counted as false positives nor as true positives in the final analysis; the same approach was used by (\cite{Ginn10a,Jaco15a}). Irrelevant findings were excluded in the evaluation because they  constitute pulmonary abnormalities that
could be important for different clinical diagnosis (\cite{Arma11}). As such, a CAD mark on such a lesion
is not a true false positive mark. It also alleviates the problem of disagreement as to what constitutes a nodule (\cite{Arma09,Ginn10a}).

\section{LUNA16 challenge}

The proposed evaluation framework was coined the LUng Nodule Analysis 2016 (LUNA16) challenge. LUNA16 invites participants to develop a CAD system that automatically detects pulmonary nodules in CT scans. The challenge provides the data set and the reference annotations described in Section~\ref{sec:data}. This data set can be used for training of the systems and the evaluation of the algorithms is performed on the same data set. This makes LUNA16 a completely open challenge. To prevent biased results as a result of training and testing on the same data set, participants are instructed to perform cross-validation in the manner described in the following subsections. The LUNA16 website allows participants to submit the results. Submitted results are automatically evaluated and presented on the website.

\subsection{Challenge tracks}
The challenge consists of two separate tracks: (1) complete nodule detection and (2) false positive reduction. 

The complete nodule detection track requires the participants to develop a complete CAD system, meaning that the only input into the system is a CT scan. 

In the false positive reduction track, participants are only required to classify a number of locations in each scan as being a nodule or not. This is equivalent to the so-called false positive reduction step in many published CAD systems. For this track, a list of nodule candidates computed using existing nodule candidates detection algorithms is supplied to the participants (see Section~\ref{sec:candidate_detectors}). This can be seen as a typical machine learning task, where a two class classification (nodule/not-nodule) has to be performed. We included this track in the challenge to encourage the participation of teams with experience in image classification tasks but no particular background on the analysis of medical images. As further support, we included a tutorial on the LUNA16 website on how to extract cubes and patches around the nodule candidate locations in CT scans.

\subsection{Cross-validation}
Participants are required to perform 10-fold cross-validation when they use the provided LIDC-IDRI data both as training and as test data. The data set has been randomly split into ten subsets of equal size on a patient level. The subsets can be directly downloaded from the LUNA16 website. 
The following steps describe how to perform 10-fold cross-validation (for fold N):
\begin{enumerate}
\item Split the data set into a test set and a training set (Subset N is used as test set and the remaining folds are used as the training set).
\item For the 'false positive reduction' track, test and training candidates should be extracted on the corresponding test and training set.
\item Train the algorithm on the training set.
\item Test the trained algorithm on the test set and generate the result file.
\item After iterating this process over all folds, merge the result files to get the result for all cases.
\end{enumerate}

\subsection{Evaluation}
\label{sec:eval}

The results of the algorithms must be submitted online in the form of a comma separated value (csv) file. The csv file contains all marks produced by the CAD system. For each CAD mark, a position (image identifier, x, y, and z coordinates) and a score should be provided. The higher the score, the more likely the location is a true nodule. 

A CAD mark is considered a true positive if it is located within a distance $r$ from the center of any nodule included in the reference standard, where $r$ is set to the radius of the reference nodule. When a nodule is detected by multiple CAD marks, the CAD mark with the highest score is selected. CAD marks that detect irrelevant findings are discarded from the analysis and are not considered as either false positive or true positive. CAD marks not falling into previous categories are marked as false positives.

Results are evaluated using the Free-Response Receiver Operating Characteristic (FROC) analysis (\cite{Rece08}). The sensitivity is defined as the fraction of detected true positives (TPs) divided by the number of nodules in our reference standard. In the FROC curve, sensitivity is plotted as a function of the average number of false positives per scan (FPs/scan). 
For each scan, we take a maximum of 100 CAD marks that were given the highest scores.
The 95\% confidence interval of the FROC curve are computed using bootstrapping with 1,000 bootstraps, as detailed in \cite{Efro94}.
In order to evaluate and compare different systems easily, we defined one overall output score. 
The overall score is defined as the average of the sensitivity at seven predefined false positive rates: $1/8$, $1/4$, $1/2$, $1$, $2$, $4$, and $8$ FPs per scan. The performance metric was introduced in the ANODE09 challenge and is referred to as the Competition Performance Metric (CPM) in \cite{Niem11a}.

The evaluation script is publicly available on the LUNA16 website and can thus be viewed and used by all participants.

\section{Methods}
\label{sec:methods}
In this section we provide a brief description of the algorithms applied in the LUNA16 challenge. As of 31 October 2016, seven systems have been applied to the complete nodule detection track and five systems have been applied to the false positive reduction track. 
The candidate detection algorithms that were used to generate candidates for false positive reduction track are presented in Section~\ref{sec:candidate_detectors};
the systems submitted to the complete detection system track are described in Section~\ref{sec:complete_cad_systems}; systems submitted to the false positive reduction track are detailed in Section~\ref{sec:false_positive_reduction}.

\subsection{Candidate detection}
\label{sec:candidate_detectors}

All candidate detection algorithms were developed as part of published CAD systems (\cite{Murp09,Jaco14,Seti15a,Tan11e,Torr15}), some of which are included in the complete nodule detection track. 
As candidates from multiple algorithms are likely to be complementary, we merged all candidates using the procedure described in Section~\ref{sec:combining_systems_cand}.
The list of merged candidates can be downloaded from the LUNA16 website and can be used by teams that want to participate in the false positive reduction track.

\subsubsection{ISICAD}
\label{isiCAD}
The generic nodule candidate detection algorithm was developed by \citet{Murp09}. 
First, the image is downsampled from $512\times512$ to $256\times256$ with the number of slices reduced to form isotropic resolution.
Thereafter, Shape Index (SI), and curvedness (CV) are computed at every voxel in the lung volume as follows:
$$ SI = \frac{2}{\pi} \arctan(\frac{k_{1} + k_{2}}{k_{1}-k_{2}}) $$
$$ CV = \sqrt[2]{k_{1}^{2}+k_{2}^{2}} $$
where $k_{1}$ and $k_{2}$ are principal curvatures computed using first and second order derivatives of the image with a Gaussian blur of scale $\sigma=1$ voxel.
After SI and CV are computed, thresholding on these values is applied to obtain seed points for nodule candidates. These seed points represent voxels which may lie on a nodule surface. Seeds are expanded using broader thresholds to form voxel clusters. To reduce the number of the clusters, clusters within 3~voxels are merged recursively. The center of the mass of the cluster is considered to be the point of interest.
The algorithm was developed using a data set from a large European lung cancer screening trial.

\subsubsection{SubsolidCAD}
\label{subsolidCAD}
The candidate detection algorithm was built to detect subsolid nodules, which are less common but more likely to be cancerous (\cite{Hens02}). 
The candidate detection algorithm by \cite{Jaco14} applies a double threshold density mask. The HU values commonly observed in subsolid nodules are used, ranging between -750~HU and -300~HU. 
Since partial volume effects may occur at the boundaries of the lungs, vessels, and airways, a morphological opening using spherical structuring element (3 voxels diameter) is applied to remove these structures. Next, connected component analysis is performed. Components with a volume smaller than 34 mm$^{3}$ are discarded from the list of candidates as subsolid nodules with a diameter smaller than 5~mm do not require follow-up CT. The centers of the candidate regions are used as nodule candidate locations.
The algorithm was developed using a data set from a large European lung cancer screening trial.

\subsubsection{LargeCAD}
\label{largeCAD}
The algorithm has the function of detecting large nodules (\cite{Seti15a}). Large solid nodules ($\geq$ 10 mm) have surface/shape index values or specific intensity range that is not captured by the two previously described nodule detection algorithms. 
An intensity threshold of -300~HU (usually corresponding to solid nodules) is applied in combination with multiple morphological operations. 
Thereafter, all connected voxels are clustered using connected component analysis; clusters with an equivalent diameter outside the range [8,40] mm are discarded.
The algorithm was developed using the data set used by LUNA16.

\subsubsection{ETROCAD}
\label{cand:etrocad}

The applied method uses the detector system proposed by \cite{Tan11e}.
Isotropic re-sampling of the image to a voxel dimension of 1~mm$^3$ is applied in the preprocessing step. 
The nodule candidate algorithm consists of a nodule segmentation method based on nodule and vessel enhancement filters and a computed divergence feature to locate the centers of the nodule clusters.
Three different set of filters (\cite{Li03e,li2004selective}) are applied to detect different types of nodules: isolated, juxtavascular, and juxtapleural nodules. 
To better estimate the location of the nodule centers and reduce the FP rate, the maxima of the divergence of the normalized gradient (DNG) of the image $k=\nabla(\overrightarrow{w})$ is used, where $\overrightarrow{w} = \frac{\overrightarrow{\Delta}L}{|| \overrightarrow{\Delta}L ||}$ and $L$ is the image intensity.
The enhancement filters and DNG are calculated at different scales in order to detect the seed points for different sizes of nodules.

Thresholding on the filtered image and DNG is applied to obtain the list of candidates.
Different thresholds on the filtered image and the nodule-enhanced image are applied for isolated nodules, juxtavascular nodules, and juxtapleural nodules to get candidate locations. Finally, to ensure that a single nodule is represented by a single mark, cluster merging is performed.
The algorithm was developed using a set of scans from LIDC-IDRI.

\subsubsection{M5L}
\label{M5L}
The candidate detection algorithm proposed by \citet{Torr15} consists of two different algorithms: LungCAM and Voxel-Based Neural Approach (VBNA). 

LungCAM is inspired based on the life-cycle of ants colonies (\cite{Cere10}). 
The lung internal structures are segmented by iteratively deploying ant colonies in voxels with intensity above a predefined thresholds. 
The ant colony moves to a specific destination and releases pheromones based on a set of rules (\cite{chialvo1995swarms}).
Voxels visited by ant colonies are removed and new ant colonies are deployed in not-yet-visited voxels.
Iterative thresholding of the pheromone maps is applied to obtain a list of candidates. 
The probability $P_{ij}$ that a candidate destination is chosen is defined as:
$$ P_{ij}(v_{i} \rightarrow v_{j}) = \frac{W(\sigma_{j})}{\sum_{n=1,26} W(\sigma_{n})} $$
where $W(\sigma_{j})$ depends on the amount of pheromone in voxel $v_{j}$.
The algorithm ends when all the ants in the colony have died.

VBNA uses two different procedures to detect nodules inside the lung parenchyma  (\cite{Li03e,Reti08}) and nodules attached to the
pleura (\cite{Reti09a}). 
The nodules inside the lung parenchyma are detected using a dedicated dot-enhancement filter. Since nodules can manifest with a different size range, a multi-scale approach is followed (\cite{Li03e}). Nodule candidate locations are defined as the local maxima of the filtered image.
The pleural nodules are detected by computing the surface normal at the lung wall. 
To build the normal, a marching cube algorithm is used. 
For each voxel inside the lung, the number of surface normals passing through are accumulated.
Pleural candidates are defined as the local maxima of the accumulated scores.
The algorithm was developed using a set of scans from LIDC-IDRI, ANODE09, and ITALUNG-CT.

\subsubsection{Combining candidate detection algorithms}
\label{sec:combining_systems_cand}

The combination of different CAD systems has been shown to improve the overall detection performance for nodule detection in chest CT (\cite{Ginn10a,Niem11a}). 
The previously described candidate detection algorithms used different approaches to detect nodules and are therefore likely to detect different sets of nodules. 
Consequently, the combination of multiple algorithms may improve the detection sensitivity of nodules and would therefore be a better baseline for the false positive reduction systems.

To combine the results of multiple candidate detection algorithms, we concatenated the lists of candidates, where candidates located closer than 5 mm to each others were merged. 
The position of the merged candidates were averaged.
Candidates located outside the lung region were discarded, as they were irrelevant for nodule detection. The lung region is determined based on the lung segmentation algorithm proposed by \cite{Rikx09b}. As the algorithm may exclude nodules attached to the lung wall, a slack border of 10~mm was applied.

\subsection{Complete nodule detection system}

The seven methods that were submitted to the complete nodule detection track are described in this section.

\label{sec:complete_cad_systems}

\subsubsection{ZNET}
ZNET uses ConvNets for both candidate detection and false positive reduction. As a preprocessing step, CT images are resampled to isotropic resolution of 0.5~mm. 
Candidate detection is extracted based on the probability map given by U-Net (\cite{Ronn15}). U-net is applied on each axial slice. Before applying U-Net, the resampled input slice is cropped to $512\times512$. 
The candidates are extracted based on the slice-based probability map output of the U-net. A threholding is applied to obtain candidate masks. The threshold was determined on the validation subset, maximizing the number of detected nodules. Thereafter, a morphological erosion operation with a 4-neighborhood kernel is used to remove partial volume effects. The candidates are then grouped by performing connected component analysis. The center of mass of the components represent the coordinates of the candidates. The false positive reduction is described in Section~\ref{sec:znet}.
Both candidate detection and false positive reduction stages were trained in a cross-validation using the provided folds from LUNA16.

\subsubsection{Aidence}

Aidence is a company developing computer assisted diagnosis tools for radiologists based on deep learning (\url{http://aidence.com/}). 
The LUNAAidence algorithm uses end-to-end ConvNets trained on a subset of studies from the National Lung Screening Trial (NLST) with additional annotation provided by in-house radiologists. 
The LUNA16 data set was used for validation purposes only and was not used as training data.
A detailed description is not available because of commercial confidentiality.

\subsubsection{JianPeiCAD}

JianPeiCAD is a system developed by Hangzhou Jianpei Technology Co. Ltd., a company based on Hangzhou, China (\url{http://www.jianpeicn.com}).
The algorithm follows the common two stage work-flow of nodule detection: Candidate detection and false positive reduction. A multi-scale rule-based screening is applied to obtain nodule candidates. The false positive reduction uses 3D ConvNets with wide channels, which are trained using data augmentation to prevent overfitting.
The system was developed using in-house resources (Chinese patient CT images and CT devices from local-vendors) and the LUNA16 data set was used as further validation for patients outside China. A detailed description is not available because of commercial confidentiality.

\subsubsection{MOT\_M5Lv1}

The Multi Opening and Threshold CAD is a fully automatic CAD developed to be included into the M5L system (\cite{Torr15}).
The lung volume is obtained using 3D region growing, with trachea exclusion and lung separation procedures.
The candidate detection algorithm was developed based on the method proposed by \cite{Mess10a}. Multiple gray level-thresholding and morphological processing is used to detect and segment nodule candidates. 
Several modifications to the sequence of threshold and opening radius as well as the merging procedure are made. 
Subsequently, a dedicated nodule segmentation method (\cite{Kuhn06}) is applied to separate nodules from vascular structures during the segmentation step.
The false positive reduction computes 15 features, among which geometrical (e.g.\ radius, sphericity, skewness of distance from center) and intensity features (e.g.\ average, standard deviation, maximum, entropy).
Classification is performed using feed-forward neural networks that consists of 1 input layer with 15~input units, 1 hidden layer with 31~units, and 1 output layer with 1~output unit.
The algorithm was developed using the LUNA16 data set.

\subsubsection{VISIACTLung}

This submission contains the results of the commercially available Visia\textsuperscript{TM} CT Lung CAD system, version 5.3 (MeVis Medical Solutions AG, Bremen, Germany). 
This is an FDA approved CAD system designed to assist radiologists in the detection of solid pulmonary nodules during review of multidetector CT scans of the chest. 
It is intended to be used as an adjunct, alerting the radiologist – after his or her initial reading of the scan – to regions of interest (ROIs) that may have been initially overlooked. A detailed description is not available because of commercial confidentiality.

\subsubsection{ETROCAD}

ETROCAD is a CAD system adapted from \citet{Tan11e}. The candidate detection algorithm is described in Section~\ref{cand:etrocad}. The false positive reduction stage uses a dedicated feature extraction and classification algorithm. For each candidate, a set of features is computed, including invariant features defined on a 3D gauge coordinates system~(\cite{Sald91}), shape features, and regional features. The classification is performed using a SVM classifier with a radial basis function.
The algorithm was developed using a set of scans from LIDC-IDRI.

\subsubsection{M5LCAD}

M5LCAD is a CAD system developed by \citet{Torr15}, which consists of two algorithms: LungCAM and VBNA. This CAD system uses the candidate detector algorithms described in section \ref{M5L}. 
The false positive reduction stage of LungCAM computes a set of 13 features for nodule candidate analysis, including spatial, intensity, and shape features. The set of features is used to classify the candidates using a feed-forward artificial neural network (FFNN). The FFNN architecture consists of 13 input neurons, 1 hidden layer with 25 neurons, and 1 neuron as output layer.
The false positive reduction of VBNA performs the classification using a standard three-layered FFNN using the raw voxels as the feature vector (\cite{Reti08,Reti09a}).
The algorithm was developed using a set of scans from LIDC-IDRI, ANODE09, and ITALUNG-CT.

\subsection{False positive reduction systems}

The five methods that were applied to the false positive reduction track are described in this section.

\label{sec:false_positive_reduction}
\subsubsection{CUMedVis}
\label{sec:CUMedVis}
CUMedVis uses multi-level contextual 3D ConvNets developed by \cite{Dou16b}.
To tackle challenges coming from variations of nodule sizes, types, and geometry characteristics, a system that consists of three different 3D~ConvNets architectures (\textit{Archi-a}, \textit{Archi-b}, \textit{Archi-c}) was presented. Each subsystem uses an input image with different receptive field so that multiple levels of contextual information surrounding the suspicious location could be incorporated.

\textit{Archi-a} has a receptive field of $20\times20\times6$. 
Three convolutional layers are used with 64 kernels of $5\times5\times3$,  $5\times5\times3$, $5\times5\times1$, respectively. 
Thereafter, a fully-connected layer with 150 output units and a softmax layer are applied.
\textit{Archi-b} has a receptive field of $30\times30\times10$. 
The first convolutional layer with 64 kernels of $5\times5\times3$ is used followed by a max-pooling layer with kernel $2\times2\times1$. 
Two convolutional layers each with 64 kernels of $5\times5\times3$ are then added, finalized by a fully-connected layer with 250 output units and a softmax layer.
\textit{Archi-c} has the largest receptive field of $40\times40\times26$. 
After the first convolutional layer with 64 kernels of $5\times5\times3$, a max-pooling layer with kernel $2\times2\times2$ is used. 
Thereafter, two convolutional layers each with 64 kernels of $5\times5\times3$ are added. 
Finally, a fully-connected layer with 250 output units and a softmax layer are established.
The prediction probabilities from the three ConvNets architectures are fused with weighted linear combination to produce the final prediction for a given candidate.

For pre-processing, voxel intensities are clipped into the interval from -1000 to 400~HU and normalized into the range of 0 to 1. To deal with the class imbalance between the false positives and nodules, translation (one voxel along each axis) and rotation (90$^0$, 180$^0$, 270$^0$ within the transverse plane) augmentations are performed on the nodules. The weights are initialized using a Gaussian distribution and are optimized using the standard back-propagation with momentum (\cite{sutskever2013importance}). A dropout strategy (\cite{hinton2012improving}) was applied during training. The system was implemented using Theano (\cite{bastien2012theano}) and a GPU of NVIDIA TITAN Z was used for acceleration.
The algorithm was developed using the cerebral microbleeds data set and was further optimized using LUNA16 data set.

\subsubsection{JackFPR}

The proposed method uses a similar multi-level contextual 3D~ConvNet architecture as presented by \cite{Dou16b}. 
It uses the three architectures (\textit{Archi-a}, \textit{Archi-b}, \textit{Archi-c}) described in Section~\ref{sec:CUMedVis} with several modifications. Exponential activation units were used as the activation functions. So instead of combining the predictions of three ConvNets using linear combination, the fully-connected layers from three architectures were concatenated and connected to a fully-connected layer with 128 output units. The last fully-connected layer was then followed by a softmax layer to obtain the prediction.

The training was performed for 240 epochs with 1,024 iterations per epoch.
Xavier initialization (\cite{glorot2010understanding}) was used as the weight initialization and Nesterov accelerated Stochastic Gradient Descent (SGD) was used. Cross-entropy loss, L2 regularization loss, and center loss were used as the cost function. 
Center loss penalizes the difference between a running average of learned features for each class and sample class features seen during the particular batch (\cite{Wen16}).
The learning rate was set to 0.005 for the first 5 epochs as a warming up. Thereafter, the learning rate was set to 0.01 and was reduced by 1/10 every 80 epochs.
Data augmentation was performed and dropout was applied to combat over-fitting.
The algorithm was developed using LUNA16 data set.

\subsubsection{DIAG CONVNET}
This method uses multi-view ConvNets proposed by \citet{Seti16}. 
For each candidate, nine $65\times65$ patches of $50\times50$~mm from different views are extracted. Each view corresponds to a different plane of symmetry in a cube and is processed using a stream of 2D~ConvNets.
The ConvNets stream consists of 3 consecutive convolutional layers and max-pooling layers:
The first is formed by 24 kernels of $5\times5$; the second by 32 kernels of $3\times3$; and the third by 48 kernels of $3\times3$. 
Weights are initialized randomly and updated during training. The max-pooling layer is used to reduce the size of patches by half. 
The last layer is a fully connected layer with 16 output units. Rectified linear units (ReLU) are used as the activation functions.
The fusion of the different ConvNets is performed using the late fusion method (\cite{Pras13,Karp14}). 
Fully-connected layers from all streams are concatenated and are connected directly to a softmax layer. 
This approach allows the network to learn 3D characteristics by comparing outputs from multiple ConvNets streams. In this approach, all the parameters of the convolutional layers from different streams are shared.

Data augmentation was applied on candidates in the training set to increase the variance of presentable candidates. For each candidate, random zooming $[0.9,1.1]$ and random rotation $[{-20}^\circ,+20^\circ]$ were performed. To prevent over-fitting during training, random positive and negative candidates with equal distribution were sampled in a batch of 64 samples.
Validation was performed every 1,024 batches.
Training was stopped when the area under the curve of the receiver operating characteristic on the validation data set does not improve after 3 epochs.
Xavier initialization (\cite{glorot2010understanding}) was used as the weight initialization.
The weights were optimized using RMSProp (\cite{Tiel12}), and evaluation was performed in a 10-fold cross validation.
Compared to the original work (\cite{Seti16}), the submitted system uses an ensemble of three multi-view ConvNets trained using different random seed-points, averaging out biases from training using random samples.
The system was implemented using Theano (\cite{bastien2012theano}); a NVIDIA TITAN X GPU was used for acceleration.
Three different architectures were evaluated by \cite{Seti16} using the same LUNA16 data set and the best performing architecture was selected. The algorithm was further optimized using the same LUNA16 data set.

\subsubsection{ZNET}
\label{sec:znet}
ZNET used the recently published wide residual networks (\cite{zagoruyko2016wide}). For each candidate, $64\times64$ patches from the axial, sagittal and coronal views were extracted. Each patch was processed separately by the wide residual networks. The predicted output values of the network for these three different patches were averaged to obtain the final prediction. 
The architecture used 4 sets of consecutive convolutional layers.
The first set consisted of 1 convolutional layer with 16 kernels of $3\times3$.
Sets two to four consisted of 10 convolutional layers with a stride of two, each with 96 kernels of $3\times3$, 192 kernels of $3\times3$, and 384 kernels of $3\times3$, respectively.
Each set also had a $1\times1\times{N}$ projection convolution in their skip connection, where $N$ is the number of kernels in the corresponding set.
The fourth layer was additionally connected to a global average pooling layer, resulting in a $1\times1\times384$ output image connected to the softmax layer.

Xavier initialization was used for weight initialization (\cite{glorot2010understanding}) and ADAM was used as the optimization method (\cite{xu2015show}). Leaky Rectified Linear Units were used as nonlinearities throughout the network.
Data augmentation (flipping, rotation, zooming and translation) was applied not only to the training data set but also the test data set in order to improve the test set scores.
The learning rate was reduced over time: Learning rate is decreased by 90\% after epochs 80 and 125.
All convolutional networks were implemented using the Lasagne and Theano libraries (\cite{dieleman2015lasagne,bastien2012theano}). The training was performed on a computer cluster using large range of CUDA enabled graphics cards including the Tesla K40M, Titan X, GTX 980, GTX 970, GTX 760 and the GTX 950M.
The algorithm was developed using the LUNA16 data set.

\subsubsection{CADIMI}
This method used multi-slice ConvNets. For each axial, sagittal, and coronal view, three patches are extracted at three locations: The plane in the exact candidate location and the planes 2~mm in both directions of the remaining free axes (x, y, and z). The patches were concatenated as three-dimensional arrays, which resulted in patches of $52\times52\times3$~mm centered around the candidate location.
The network consisted of 2D ConvNets with three consecutive convolutional layers and max-pooling: 
The first convolutional layer used 24 kernels of $5\times5$; the second used 32 kernels of $3\times3$; and the third 48 kernels of $3\times3$.
The output of the last max-pooling was connected to fully-connected layer of 512 output units.
ReLU was used as the activation function, and the last fully-connected layer was connected to a softmax layer. 

Training was performed one time using patches from all three views for 80 epochs. For each epoch, all positive patches and 20,000 random negative patches were used.
In order to tackle the problem of data imbalance, data augmentation (vertical/horizontal flip and random cropping) was applied.
During testing, 5 patches (1 center patch and 4 patches with $[-4,+4]$ translation in two axes) were extracted from each view. These patches were processed using a single trained network and the predictions were averaged. Batch normalization was applied after each max-pooling layer to reduce over-fitting.
The weights were initialized using the uniform initialization (\cite{he2015delving}). Nesterov accelerated SGD with a learning rate of 0.01, a decay of 0.001, and a momentum of 0.9 is used.
The system was implemented using the Lasagne and Theano libraries (\cite{dieleman2015lasagne,bastien2012theano}). 
The algorithm was developed using the LUNA16 data set.

\subsection{Combining false positive reduction systems}
\label{sec:combining_systems}

The combination of multiple classification methods, also known as an ensemble method, has been used in many machine-learning problems to improve the prediction performance (\cite{Diet00}).
As systems applied in the false positive reduction track use the same set of candidates, the impact of combining multiple methods could be evaluated.
In this study, we combined CAD results from the systems in the false positive reduction track. The combination is performed by simply averaging the probabilities given by the systems. Such is a common approach in optimizing the performance of deep learning architectures (\cite{Szeg14,He15b}). 

\subsection{Observer study}

To evaluate the potential of CAD systems to detect nodules missed by human readers, and to elucidate the nature of the false positives detected by the CAD systems, an observer study was performed. 
In the observer study, CAD marks from the combination of false positive reduction systems were assessed to identify if there were additional nodules detected. 
The reading process was performed by four expert readers independently.

We extracted all CAD marks at 0.25~FPs/scan that were categorized as false positives to be further analyzed by expert readers. 
To reduce the readers' workload, research scientists read and removed CAD marks that were obvious false positives (e.g.\ vessels, ribs, diaphragm) beforehand. 
Thereafter, CAD marks that were close to annotated lesions in LIDC-IDRI but were missed by our hit criteria (thus considered as false positives) were discarded. Most of these lesions were non-nodular and therefore were not well captured by the defined hit criteria (radius of the corresponding lesion).
This operation resulted in a set of 127 marks that were potentially nodules. As a similar observer study was performed in our previous study (\cite{Jaco15a}), marks which were already evaluated on this CT data by radiologists were not read again and the scores of the four radiologists from the previous study were used. Last, we asked the expert readers to review and annotate the remaining marks as: nodule~$\geq$~3~mm, nodule~$<$~3~mm, or false positives. Measurement tools were made available to readers during the process in order to enable size evaluation.

\section{Results}
\label{sec:results}
In this section, we present the results achieved by all individual systems described in Section \ref{sec:methods}. The results of combining multiple algorithms are provided.

\subsection{Candidate detection}

\begin{table*}
\scriptsize
\centering
\begin{tabular}{lcrrrrr}
\toprule
System name & Combination & Sensitivity & \begin{tabular}[c]{@{}c@{}}Best single\\ sensitivity\end{tabular} & \begin{tabular}[c]{@{}c@{}}Difference\\ sensitivity\end{tabular} & \begin{tabular}[c]{@{}c@{}}Total number\\ of candidates\end{tabular} & \begin{tabular}[c]{@{}c@{}}Average number of\\candidates / scan\end{tabular} \\ \midrule
ISICAD       & $\blacksquare \Box \Box \Box \Box$ & 0.856 & & & 298 256 & 335.9 \\
SubsolidCAD  & $\Box \blacksquare \Box \Box \Box$ & 0.361 & & & 258 075 & 290.6 \\
LargeCAD    & $\Box \Box \blacksquare \Box \Box$ & 0.318 & & &  42 281 & 47.6 \\
M5L          & $\Box \Box \Box \blacksquare \Box$ & 0.768 & & &  19 687 & 22.2 \\
ETROCAD      & $\Box \Box \Box \Box \blacksquare$ & 0.929 & & & 295 686 & 333.0 \\
& $ \blacksquare \blacksquare \square \square \square $                & 0.918 & 0.857 & 0.062 & 520 319 & 585.9 \\
& $ \blacksquare \square \blacksquare \square \square $                & 0.898 & 0.857 & 0.041 & 328 742 & 370.2 \\
& $ \blacksquare \square \square \blacksquare \square $                & 0.917 & 0.857 & 0.061 & 308 047 & 346.9 \\
& $ \blacksquare \square \square \square \blacksquare $                & 0.959 & 0.929 & 0.030 & 524 108 & 590.2 \\
& $ \square \blacksquare \blacksquare \square \square $                & 0.523 & 0.361 & 0.162 & 295 476 & 332.7 \\
& $ \square \blacksquare \square \blacksquare \square $                & 0.869 & 0.768 & 0.101 & 274 900 & 309.6 \\
& $ \square \blacksquare \square \square \blacksquare $                & 0.954 & 0.929 & 0.024 & 518 058 & 583.4 \\
& $ \square \square \blacksquare \blacksquare \square $                & 0.834 & 0.768 & 0.066 &  59 359 &  66.8 \\
& $ \square \square \blacksquare \square \blacksquare $                & 0.945 & 0.929 & 0.016 & 319 405 & 359.7 \\
& $ \square \square \square \blacksquare \blacksquare $                & 0.942 & 0.929 & 0.013 & 297 030 & 334.5 \\
& $ \blacksquare \blacksquare \blacksquare \square \square $           & 0.944 & 0.857 & 0.088 & 551 065 & 620.6 \\
& $ \blacksquare \blacksquare \square \blacksquare \square $           & 0.954 & 0.857 & 0.098 & 530 942 & 597.9 \\
& $ \blacksquare \blacksquare \square \square \blacksquare $           & 0.977 & 0.929 & 0.048 & 728 162 & 820.0 \\
& $ \blacksquare \square \blacksquare \blacksquare \square $           & 0.934 & 0.857 & 0.078 & 339 229 & 382.0 \\
& $ \blacksquare \square \blacksquare \square \blacksquare $           & 0.964 & 0.929 & 0.035 & 548 523 & 617.7 \\
& $ \blacksquare \square \square \blacksquare \blacksquare $           & 0.967 & 0.929 & 0.038 & 529 404 & 596.2 \\
& $ \square \blacksquare \blacksquare \blacksquare \square $           & 0.900 & 0.768 & 0.132 & 310 323 & 349.5 \\
& $ \square \blacksquare \blacksquare \square \blacksquare $           & 0.964 & 0.929 & 0.035 & 545 204 & 614.0 \\
& $ \square \blacksquare \square \blacksquare \blacksquare $           & 0.965 & 0.929 & 0.035 & 524 726 & 590.9 \\
& $ \square \square \blacksquare \blacksquare \blacksquare $           & 0.954 & 0.929 & 0.024 & 326 274 & 367.4 \\
& $ \blacksquare \blacksquare \blacksquare \blacksquare \square $      & 0.980 & 0.929 & 0.051 & 750 838 & 845.5 \\
& $ \blacksquare \blacksquare \blacksquare \square \blacksquare $      & 0.983 & 0.929 & 0.054 & 732 901 & 825.3 \\
& $ \blacksquare \blacksquare \square \blacksquare \blacksquare $      & 0.970 & 0.929 & 0.040 & 553 327 & 623.1 \\
& $ \blacksquare \square \blacksquare \blacksquare \blacksquare $      & 0.965 & 0.857 & 0.108 & 559 543 & 630.1 \\
& $ \square \blacksquare \blacksquare \blacksquare \blacksquare $      & 0.970 & 0.929 & 0.040 & 551 227 & 620.8 \\
& $ \blacksquare \blacksquare \blacksquare \blacksquare \blacksquare $ & 0.983 & 0.929 & 0.054 & 754 975 & 850.2 \\

\bottomrule
\end{tabular}
\caption{The results of five candidate detection systems and all possible combinations. The filled squares indicate which systems were included in the combination. CPM: Competition Performance Metric.}
\label{table:3}
\end{table*}

Table~\ref{table:3} summarizes the performance of individual candidate detection algorithms and their top performing combinations. The best detection sensitivity of 92.9\% was achieved by ETROCAD. 
When multiple candidate detection algorithms were combined, the sensitivity substantially improved up to 98.3\% (1,166/1,186 nodules), higher than the sensitivity of any individual system. This illustrates the potential of combining multiple candidate detection algorithms to improve the sensitivity of CAD systems.

\subsection{Complete nodule detection track}

The FROC curves of the systems on the complete nodule detection track are shown in Figure~\ref{fig:FROC_NDET}. 
In this track, the best score was achieved by ZNET with a CPM of 0.811. 
Other systems show comparable performance.
It was observed that the relatively large differences in terms of sensitivity at low FPs/scan substantially influences the overall scores of the systems.

\subsection{False positive reduction track}

\begin{figure*}[t]
\centering
\begin{subfigure}[t]{0.48\textwidth}
  \includegraphics[width=1.0\linewidth]{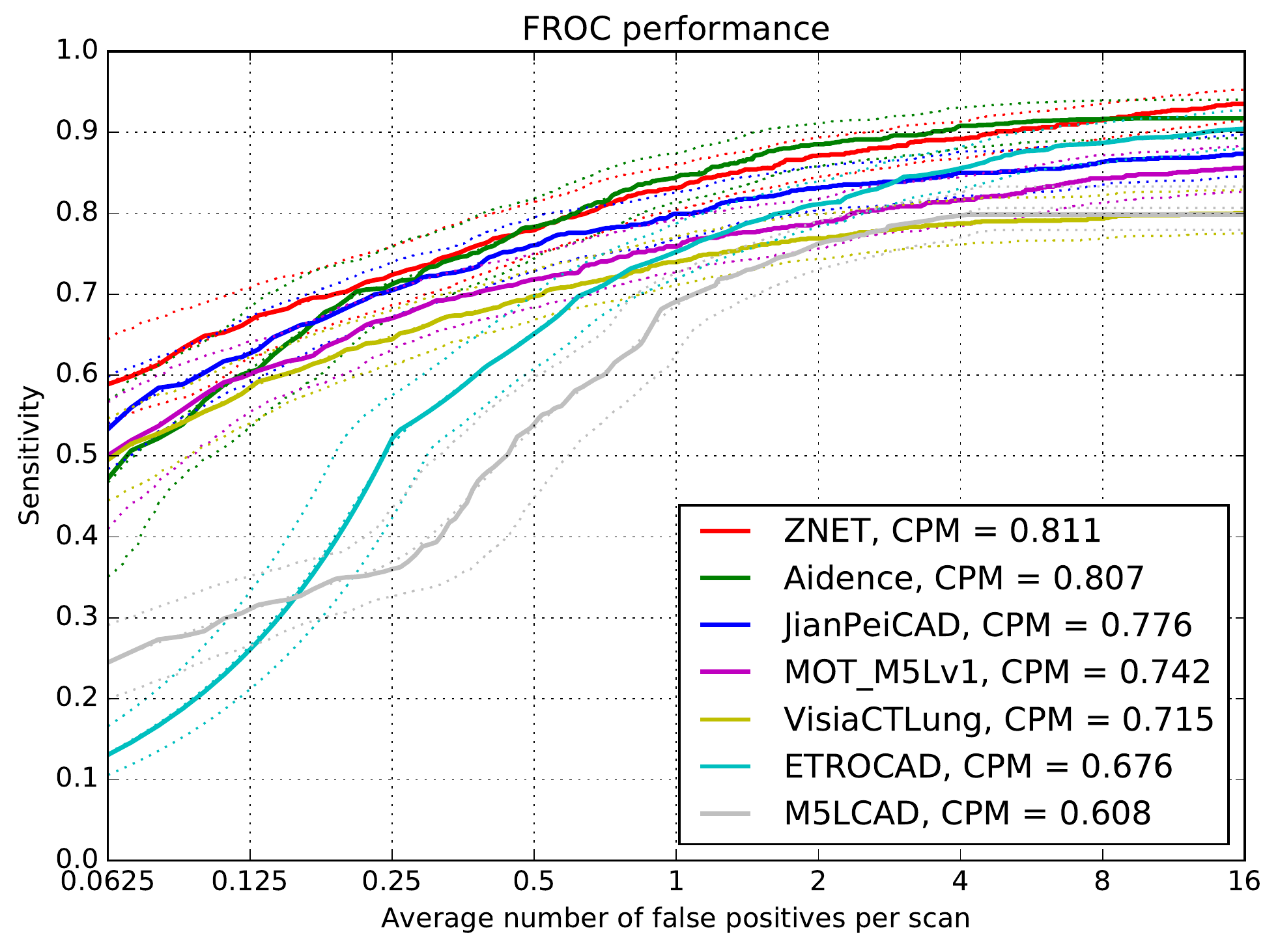}
  \subcaption{\label{fig:FROC_NDET}}
\end{subfigure}
~
\begin{subfigure}[t]{0.48\textwidth}
  \includegraphics[width=1.0\linewidth]{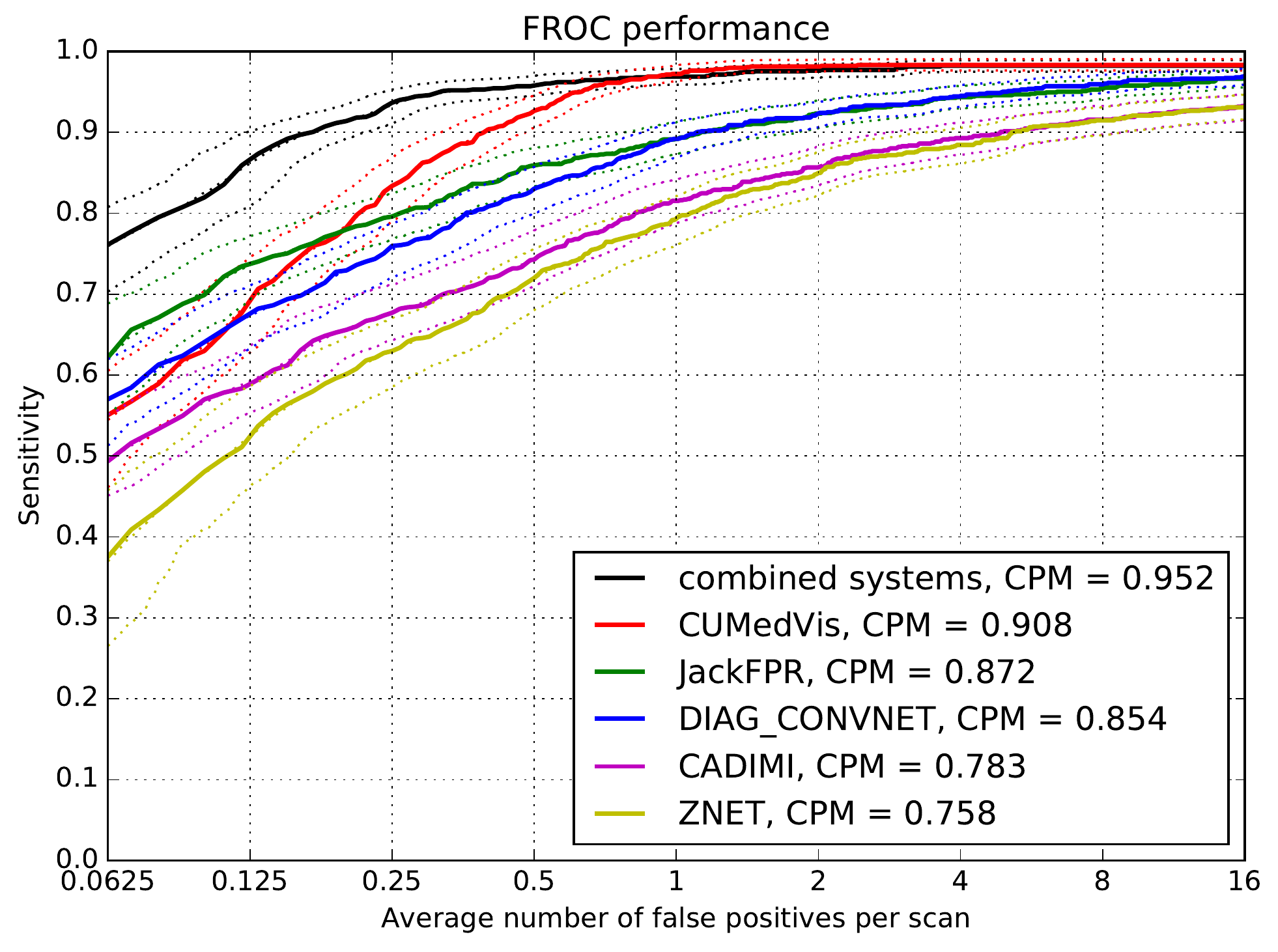}
  \subcaption{\label{fig:FROC_FPRED}}
\end{subfigure}

\caption{FROC curves of the systems in (a) the nodule detection track and (b) the false positive reduction track. Dashed curves represent the 95\% confidence intervals estimated using bootstrapping.
\label{fig:FROC_LUNA16}}
\end{figure*}

The FROC curves of the systems on the false positive reduction track are shown in Figure~\ref{fig:FROC_FPRED}. The best average score was achieved by CuMedVis, with a CPM of 0.908. 
Table~\ref{tab:CPM} shows all possible system combinations, where the sensitivities of the combined systems were higher than the sensitivity achieved by the best system.
Although all false positive reduction systems are based on ConvNets, it is evident that combining ConvNets with different configurations further improves the overall sensitivity as shown in Table~\ref{tab:CPM}. The p-value was defined as the probability that a system's CPM is higher or lower than the reference system's CPM. A p-value below 0.002 is considered to be statistically significant after Bonferroni correction ($m=30$).

\begin{table*}
\scriptsize
\centering
\begin{tabular}{@{}lcrrrrrrrrrrr@{}}
\hline
System name & Combination & 0.125 & 0.25 & 0.5 & 1 & 2 & 4 & 8 & CPM & \begin{tabular}[c]{@{}c@{}}P-value\end{tabular} & \begin{tabular}[c]{@{}c@{}}Difference\\CPM\end{tabular}& \begin{tabular}[c]{@{}c@{}}Best single\\CPM\end{tabular} \\ 
\hline

CUMedVis     & $ \blacksquare \square \square \square \square $                      & 0.677 & 0.834 & 0.927 & 0.972 & 0.981 & 0.983 & 0.983 & 0.908 & Reference &       &        \\
JackFPR      & $ \square \blacksquare \square \square \square $                      & 0.734 & 0.796 & 0.859 & 0.892 & 0.923 & 0.944 & 0.954 & 0.872 &    0.002  &       &        \\
DIAG CONVNET & $ \square \square \blacksquare \square \square $                      & 0.669 & 0.760 & 0.831 & 0.892 & 0.923 & 0.945 & 0.960 & 0.854 & $<$0.001  &       &        \\
CADIMI       & $ \square \square \square \blacksquare \square $                      & 0.583 & 0.677 & 0.743 & 0.815 & 0.857 & 0.893 & 0.916 & 0.783 & $<$0.001  &       &        \\
ZNET         & $ \square \square \square \square \blacksquare $                      & 0.511 & 0.630 & 0.720 & 0.793 & 0.850 & 0.884 & 0.915 & 0.758 & $<$0.001  &       &        \\
             & $ \blacksquare \blacksquare \square \square \square $                 & 0.809 & 0.901 & 0.962 & 0.976 & 0.981 & 0.981 & 0.982 & 0.942 & $<$0.001  & 0.908 & 0.034  \\
             & $ \blacksquare \square \blacksquare \square \square $                 & 0.831 & 0.917 & 0.965 & 0.979 & 0.981 & 0.981 & 0.981 & 0.948 & $<$0.001  & 0.908 & 0.040  \\
             & $ \blacksquare \square \square \blacksquare \square $                 & 0.802 & 0.903 & 0.948 & 0.976 & 0.979 & 0.979 & 0.980 & 0.938 & $<$0.001  & 0.908 & 0.030  \\
             & $ \blacksquare \square \square \square \blacksquare $                 & 0.831 & 0.927 & 0.968 & 0.976 & 0.979 & 0.981 & 0.981 & 0.949 & $<$0.001  & 0.908 & 0.041  \\
             & $ \square \blacksquare \blacksquare \square \square $                 & 0.745 & 0.826 & 0.864 & 0.906 & 0.948 & 0.958 & 0.969 & 0.888 &    0.042  & 0.872 & 0.016  \\
             & $ \square \blacksquare \square \blacksquare \square $                 & 0.717 & 0.797 & 0.858 & 0.895 & 0.932 & 0.947 & 0.959 & 0.872 & $<$0.001  & 0.872 & 0.000  \\
             & $ \square \blacksquare \square \square \blacksquare $                 & 0.728 & 0.828 & 0.879 & 0.917 & 0.938 & 0.954 & 0.963 & 0.887 &    0.038  & 0.872 & 0.015  \\
             & $ \square \square \blacksquare \blacksquare \square $                 & 0.550 & 0.680 & 0.796 & 0.869 & 0.912 & 0.938 & 0.959 & 0.815 & $<$0.001  & 0.854 & -0.040 \\
             & $ \square \square \blacksquare \square \blacksquare $                 & 0.616 & 0.737 & 0.831 & 0.888 & 0.931 & 0.953 & 0.964 & 0.845 & $<$0.001  & 0.854 & -0.009 \\
             & $ \square \square \square \blacksquare \blacksquare $                 & 0.602 & 0.732 & 0.812 & 0.852 & 0.884 & 0.913 & 0.946 & 0.820 & $<$0.001  & 0.783 & 0.037  \\
             & $ \blacksquare \blacksquare \blacksquare \square \square $            & 0.821 & 0.898 & 0.954 & 0.975 & 0.981 & 0.982 & 0.982 & 0.942 & $<$0.001  & 0.908 & 0.034  \\
             & $ \blacksquare \blacksquare \square \blacksquare \square $            & 0.816 & 0.897 & 0.945 & 0.970 & 0.980 & 0.980 & 0.980 & 0.938 & $<$0.001  & 0.908 & 0.030  \\
             & $ \blacksquare \blacksquare \square \square \blacksquare $            & 0.843 & 0.911 & 0.957 & 0.978 & 0.981 & 0.981 & 0.981 & 0.947 & $<$0.001  & 0.908 & 0.039  \\
             & $ \blacksquare \square \blacksquare \blacksquare \square $            & 0.817 & 0.912 & 0.954 & 0.968 & 0.975 & 0.979 & 0.982 & 0.941 & $<$0.001  & 0.908 & 0.033  \\
             & $ \blacksquare \square \blacksquare \square \blacksquare $            & 0.859 & 0.937 & 0.958 & 0.969 & 0.976 & 0.982 & 0.982 & 0.952 & $<$0.001  & 0.908 & 0.044  \\
             & $ \blacksquare \square \square \blacksquare \blacksquare $            & 0.820 & 0.907 & 0.946 & 0.968 & 0.976 & 0.981 & 0.981 & 0.940 & $<$0.001  & 0.908 & 0.032  \\
             & $ \square \blacksquare \blacksquare \blacksquare \square $            & 0.720 & 0.802 & 0.864 & 0.916 & 0.941 & 0.960 & 0.970 & 0.882 &    0.010  & 0.872 & 0.010  \\
             & $ \square \blacksquare \blacksquare \square \blacksquare $            & 0.736 & 0.835 & 0.891 & 0.924 & 0.945 & 0.969 & 0.973 & 0.896 &    0.222  & 0.872 & 0.024  \\
             & $ \square \blacksquare \square \blacksquare \blacksquare $            & 0.741 & 0.815 & 0.874 & 0.918 & 0.938 & 0.954 & 0.965 & 0.887 &    0.024  & 0.872 & 0.015  \\
             & $ \square \square \blacksquare \blacksquare \blacksquare $            & 0.635 & 0.777 & 0.839 & 0.888 & 0.929 & 0.954 & 0.965 & 0.855 & $<$0.001  & 0.854 & 0.001  \\
             & $ \blacksquare \blacksquare \blacksquare \blacksquare \square $       & 0.823 & 0.896 & 0.939 & 0.968 & 0.977 & 0.980 & 0.981 & 0.938 & $<$0.001  & 0.908 & 0.030  \\
             & $ \blacksquare \blacksquare \blacksquare \square \blacksquare $       & 0.846 & 0.912 & 0.949 & 0.971 & 0.977 & 0.981 & 0.982 & 0.946 & $<$0.001  & 0.908 & 0.037  \\
             & $ \blacksquare \blacksquare \square \blacksquare \blacksquare $       & 0.821 & 0.892 & 0.944 & 0.970 & 0.978 & 0.981 & 0.981 & 0.938 & $<$0.001  & 0.908 & 0.030  \\
             & $ \blacksquare \square \blacksquare \blacksquare \blacksquare $       & 0.830 & 0.912 & 0.947 & 0.964 & 0.973 & 0.979 & 0.981 & 0.941 & $<$0.001  & 0.908 & 0.033  \\
             & $ \square \blacksquare \blacksquare \blacksquare \blacksquare $       & 0.745 & 0.823 & 0.884 & 0.925 & 0.946 & 0.961 & 0.973 & 0.894 &    0.102  & 0.872 & 0.022  \\
             & $ \blacksquare \blacksquare \blacksquare \blacksquare \blacksquare $  & 0.836 & 0.896 & 0.940 & 0.965 & 0.976 & 0.981 & 0.982 & 0.939 & $<$0.001  & 0.908 & 0.031  \\

\bottomrule
\end{tabular}
\caption{The results of five false positive reduction systems and all possible combinations. The filled squares indicate which systems were included in the combination. CPM: Competition Performance Metric.}
\label{tab:CPM}
\end{table*}

\subsection{Performance based on nodule type}

To assess the performance of the algorithms on different types of nodules (non-solid, part-solid, and solid), additional analysis was performed. The nodule type was derived based on the morphological characteristic scored by the LIDC-IDRI radiologists. The nodule was labeled "non-solid" if the majority of the radiologists gave a texture score of 1, "solid" for a majority score of 5, and "part-solid" if the two previous criterion did not hold. Using this labelling strategy, the LUNA16 data set consisted of 64 non-solid nodules, 189 part-solid nodules, and 933 solid nodules. The performance of the algorithms on different set of nodules are tabulated in Table~\ref{tab:cpm_nodule_type}.

\begin{table*}[h]
\centering
\scriptsize
\caption{Performance benchmark of algorithms on different sets of nodules. Nodules are categorized based on the nodule type, which was derived based the morphological characteristic scored by the LIDC-IDRI radiologists (5-point scale: 1$=$non-solid, 3$=$part-solid, 5$=$solid). The nodule was labeled "non-solid" if the majority of the radiologists gave a texture score of 1, "solid" for a majority score of 5, and "part-solid" if the two previous criterion did not hold. CPM score was used as the performance metric}
\label{tab:cpm_nodule_type}

\begin{tabular}{lrrrr}
\hline
System name & all (1,186) & non-solid (64) & part-solid (189) & solid (933) \\
\hline
\textbf{Nodule detection track} & & & & \\
ZNET        & 0.811 & 0.663 & 0.735 & 0.836 \\
Aidence     & 0.807 & 0.730 & 0.776 & 0.819 \\
JianPeiCAD  & 0.776 & 0.248 & 0.725 & 0.825 \\
MOT\_M5v1   & 0.742 & 0.217 & 0.696 & 0.787 \\
VisiaCTLung & 0.715 & 0.033 & 0.652 & 0.775 \\
ETROCAD  & 0.676 & 0.290 & 0.547 & 0.728 \\
M5LCAD   & 0.608 & 0.156 & 0.549 & 0.650 \\
\textbf{False positive reduction track} & & & & \\
CUMedVis      & 0.908 & 0.908 & 0.912 & 0.907 \\
JackFPR                       & 0.872 & 0.636 & 0.845 & 0.893 \\
DIAG\_CONVNET  & 0.854 & 0.688 & 0.819 & 0.873 \\
CADIMI                       & 0.783 & 0.496 & 0.751 & 0.810 \\
ZNET                         & 0.758 & 0.498 & 0.685 & 0.790 \\

\hline

\end{tabular}
\end{table*}

\subsection{Analysis of false positives: observer study}

\begin{table}[h]
\centering
\scriptsize
\caption{An overview of the observer study on 222 false positives at 0.25~FPs/scan. The table shows the number of false positives that were accepted by the expert readers as nodule~$\geq$3~mm at different agreement levels. The number of false positives that were not accepted as nodule~$\geq$3~mm but were accepted as nodule$<$3~mm is also included.}
\label{tab:obs_study}
\begin{tabular}{lr}
\hline
Category & Number \\
\hline
nodule $\geq$3~mm - at least 1         & 108 \\
nodule $\geq$3~mm - at least 2         &  91 \\
nodule $\geq$3~mm - at least 3         &  69 \\
nodule $\geq$3~mm - at least 4         &  41 \\
nodule $<$3~mm &   6 \\
\hline

\end{tabular}
\end{table}

A summary of the observer study is shown in Table~\ref{tab:obs_study}. 
Among 127 CAD marks, 108, 91, 69, and 41 CAD marks were accepted as nodules~$\geq$~3~mm by at least 1, 2, 3, or 4 readers, respectively;
6 out of 19 remaining CAD marks were considered as nodule~$<$~3~mm.
Examples of nodules found in this observer study are shown in Figure~\ref{fig:FP_accepted}.
We shared the set of additional nodules on the LUNA16 website to be used for further development of CAD systems.

\section{Discussion}
\label{sec:discussion}

In this study, we presented LUNA16: A novel evaluation framework for automatic nodule detection algorithms. The aim of the study was to supply the research community a framework to test and compare algorithms on a common large database with a standardized evaluation protocol. This allows the community to objectively evaluate different CAD systems and push forward the development of state of the art nodule detection algorithms. The submitted systems were described and the performance was evaluated. We showed that the combination of multiple false positive reduction algorithms applied on a combined set of candidates outperformed any individual system. This highlights the potential of combining algorithms to improve the detection performance.

\begin{figure*}[t]
\centering

\includegraphics[width=0.0255\linewidth]{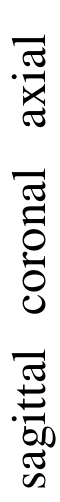}
\begin{subfigure}[t]{0.44\textwidth}
  \includegraphics[width=1.0\linewidth]{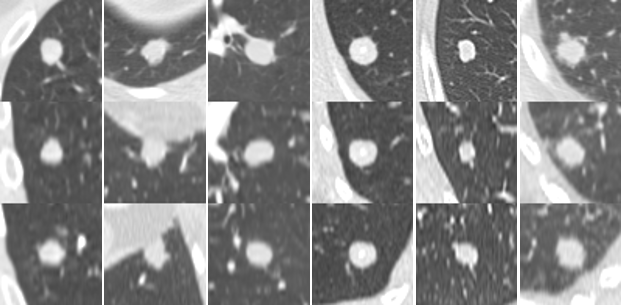}
  \subcaption{true positives with highest probability}
\end{subfigure}
\qquad~
\begin{subfigure}[t]{0.44\textwidth}
  \includegraphics[width=1.0\linewidth]{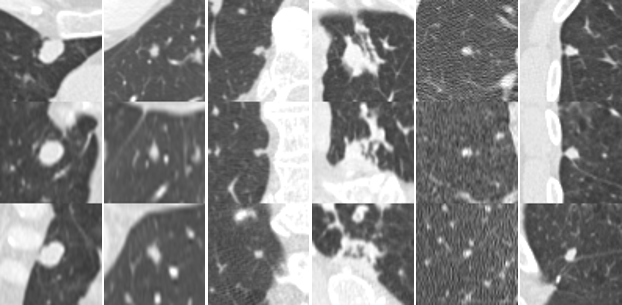}
  \subcaption{random true positives at 1~FP/scan}
\end{subfigure}

\vspace{5mm}
\includegraphics[width=0.0255\linewidth]{label.pdf}
\begin{subfigure}[t]{0.44\textwidth}
  \includegraphics[width=1.0\linewidth]{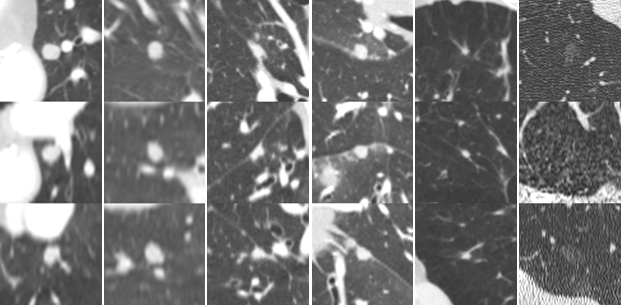}
  \subcaption{false positives accepted as nodules by radiologists \label{fig:FP_accepted}}
\end{subfigure}
\qquad~
\begin{subfigure}[t]{0.44\textwidth}
  \includegraphics[width=1.0\linewidth]{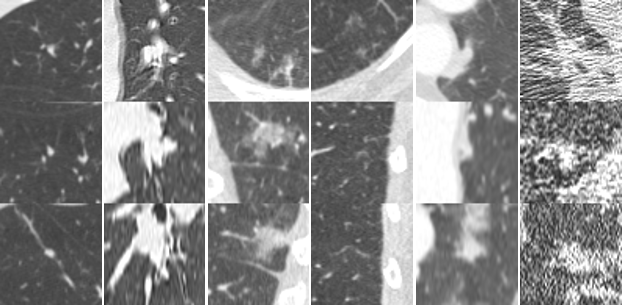}
  \subcaption{random false positives at 1~FP/scan}
\end{subfigure}

\vspace{5mm}
\includegraphics[width=0.0255\linewidth]{label.pdf}
\begin{subfigure}[t]{0.44\textwidth}
  \includegraphics[width=1.0\linewidth]{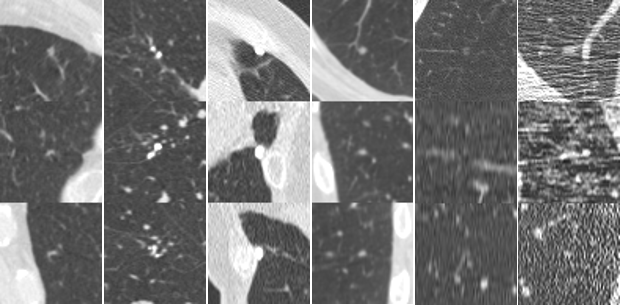}
  \subcaption{false negatives from the candidate detectors}
\end{subfigure}
\qquad~
\begin{subfigure}[t]{0.44\textwidth}
  \includegraphics[width=1.0\linewidth]{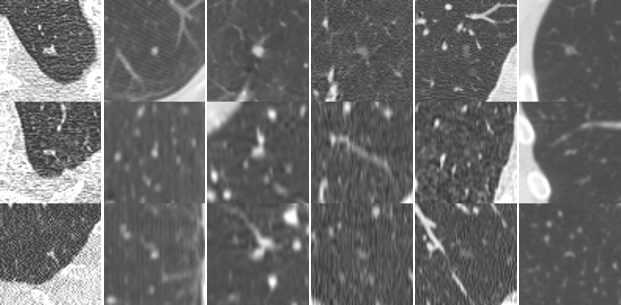}
  \subcaption{random false negatives at 1~FP/scan}
\end{subfigure}

\caption{Examples of true positives, false positives, and false negatives from the combined system. Each lesion is located at the center of the $50\times50$~mm patch in axial, coronal, and sagittal views. \label{fig:examples}}
\end{figure*}

Candidate detection plays an important role of determining the maximum attainable detection sensitivity of a CAD system. The algorithms should ideally detect all nodules with an acceptable amount of false positives.
Table~\ref{table:3} shows that the individual candidate detection algorithms achieve a detection sensitivity between 31.8\% and 92.9\%; combining different candidate detection algorithms improved the sensitivity up to 98.3\%.
While a smaller set of candidates (a combined set of only ISICAD, SubsolidCAD, and LargeCAD candidates) were also provided in the earlier phase of LUNA16, we here only reported the results of the systems that use the latest set of candidates that has a much higher sensitivity. The results of other systems that use the smaller set of candidates, resulting in lower scores, are available on the LUNA16 website. It is worth noting that the candidate detection systems used in this study do not employ deep learning, while systems in the complete nodule detection track, e.g.\ ZNET, do employ ConvNets to detect nodule candidates.

In the complete nodule detection track, a total of seven systems were evaluated. 
Diverse methods were applied and different sets of data were used for training.
When evaluated using the same data set, the detection sensitivity ranged between 69.1\% and 91.5\% at 1 and 8~FPs/scan, as shown in Figure~\ref{fig:FROC_NDET}.
Notably, the top three systems make use of ConvNets for their detection algorithms.
While the variability of the performance is determined by the underlying methods, it is also affected by the training data used to develop the system (see also Table~\ref{tab:CADListOnLIDC}).
This suggests the need of a standardized training data set for appropriate comparison of algorithms.

In the false positive reduction track, different systems for false positive reduction were evaluated given a common set of candidates and training data. A total of five systems were evaluated. ConvNets were used as the prediction model for all systems, which is in line with the recent trend of adopting deep learning in the medical image analysis domain. As shown in Figure~\ref{fig:FROC_FPRED}, all systems achieve detection sensitivity between 79.3\% and 98.3\% at 1 and 8~FPs/scan. As the underlying method is similar, one could hypothesize that there could be little to no benefit when these systems are combined. Nevertheless, combining multiple ConvNets systems did substantially improve the detection performance (black line on Figure~\ref{fig:FROC_FPRED}); a detection sensitivity of over 95.0\% was achieved at fewer than 1~FP/scan. Despite all methods being based on ConvNets, the differences in network parameters, such as selected architectures, random initialization methods, and input patches, apparently make these systems somewhat complementary for prediction, and this is leveraged by the (simple averaging) combination.

To provide a broader context to the results reported in this paper, we listed the performance of other published CAD systems that use LIDC-IDRI data in Table~\ref{tab:CADListOnLIDC}. For each CAD system, we listed the number of scans used in the validation data set, nodule inclusion criteria, the number of included nodules, and the reported CAD performance. Note that different subsets of LIDC-IDRI database were used; LUNA16 aims to make the CAD performance comparison more easy and more fair by using exactly the same data and evaluation protocol for each system. The CAD systems presented in \cite{Jaco15a} are not listed in this table as these CAD systems also participated in the LUNA16 challenge and hence are already described in this paper.

The observer study showed that some false positives detected by the CAD systems are nodules that were missed during the manual annotations of LIDC-IDRI. The majority of these nodules were overlooked because they were small or less visible (e.g.\ ground-glass/non-solid nodules).
Other nodules may be missed because there were multiple nodules in the corresponding scans, or because the nodules were part of a complex abnormality (e.g.\ an area of consolidation). While these nodules may be found during follow-up, detecting them early could provide essential clinical information (e.g.\ growth rate). 

Examples of lesions detected or missed by the combined CAD system are shown in Figure~\ref{fig:examples}. 
Nodules with a wide range of morphological characteristics are detected at 1 FP/scan, showing that ConvNets are capable of capturing morphological variation of nodules in the network.
Larger nodules are unlikely to be missed, which is just as well as there is a strong positive correlation between size and malignancy risk. 
Most false positives are large vessels, scar tissue, spinal abnormalities, and other mediastinal structures. 
These false positives are a challenge. In scans from subjects with interstitial lung disease, there are, even in mild cases, regions with irregular opacities that can lead to a large number of erroneous nodule CAD marks.
Other false positives are caused by motion artifacts and extreme noise.
The false negatives were small and/or had irregular shapes. Improving the robustness of the candidate detection algorithms to detect small nodules should further improve the performance.

\begin{table*}
\scriptsize
\renewcommand{\arraystretch}{1.3}

\caption{\label{tab:CADListOnLIDC} The performance summary of published CAD systems evaluated using LIDC-IDRI data sets. Note that different subsets of scans were used by different research groups.}
\centering
\begin{tabular}{L{0.18\linewidth} R{0.06\linewidth} R{0.06\linewidth} R{0.06\linewidth} R{0.06\linewidth} C{0.06\linewidth} R{0.06\linewidth} C{0.08\linewidth} C{0.08\linewidth}} 
\hline
CAD systems & Year & \#~scans & slice thickness & nodules size (mm) & agreement levels & \#~nodules & \multicolumn{2}{c}{sensitivity (\%) / FPs/scan} \tabularnewline
\hline
Combined LUNA16 & - & 888 & $\leq$2.5 & $\geq$3 & at least 3 & 1,186 & 98.2 / 4.0 & 96.9 / 1.0 \tabularnewline
\citet{Dou16b} & 2016 & 888 & $\leq$2.5 & $\geq$3 & at least 3 & 1,186 & 90.7 / 4.0 & 84.8 / 1.0 \tabularnewline
\citet{Seti16} & 2016 & 888 & $\leq$2.5 & $\geq$3 & at least 3 & 1,186 & 90.1 / 4.0 & 85.4 / 1.0 \tabularnewline
\citet{Berg16a} & 2016 & 243 & - & $\geq$3 & at least 1 & 690 & 85.9 / 2.5 & - \tabularnewline
\citet{Torr15} & 2015 & 949 & - & $\geq$3 & at least 2 & 1,749 & 80.0 / 8.0 & - \tabularnewline
\citet{Ginn15} & 2015 & 865 & $\leq$2.5 & $\geq$3 & at least 3 & 1,147 & 76.0 / 4.0 & 73.0 / 1.0 \tabularnewline
\citet{Brow14} & 2014 & 108 & 0.5-3 & $\geq$4 & at least 3 & 68 & 75.0 / 2.0 & - \tabularnewline
\citet{Choi13} & 2013 & 58 & 0.5-3 & 3-30 & at least 1 & 151 & 95.3 / 2.3 & - \tabularnewline
\citet{Tan13b} & 2013 & 360 & - & $\geq$3 & at least 4 & -  & 83.0 / 4.0 & - \tabularnewline 
\citet{Tera13} & 2013 & 84 & 0.5-3 & 5-20 & at least 1 & 103 & 80.0 / 4.2 & - \tabularnewline
\citet{Casc12} & 2012 & 84 & 1.25-3 & $\geq$3 & at least 1 & 148 & 97.0 / 6.1 & 88.0 / 2.5 \tabularnewline
\citet{Guo12}  & 2012 & 85 & 1.25-3 & $\geq$3 & at least 3 & 111 & 80.0 / 7.4 & 75.0 / 2.8 \tabularnewline
\citet{Cama11} & 2011 & 69	  & 0.5-2		& $>$3 & at least 2 & 114 & 80.0 / 3.0 & - \tabularnewline	 
\citet{Ricc11} & 2011 & 154  & 0.5-3		& $\geq$3 & at least 4 & 117 & 71.0 / 6.5 & 60.0 / 2.5 \tabularnewline	
\citet{Tan11e} & 2011 & 125 & 0.75-3 & $\geq$3 & at least 4 & 80 & 87.5 / 4.0 & - \tabularnewline
\citet{Mess10a} & 2010 & 84 & 1.3-3 & $\geq$3 & at least 1 & 143 & 82.7 / 3.0 & - \tabularnewline

\hline
\end{tabular}
\end{table*}

This study has several limitations. As the LIDC-IDRI is a web-accessible database for development and evaluation of CAD systems, all nodule annotations are publicly available. The usage of a completely open database is not a common setup for challenges. Typically, an independent test set is provided, for which the reference annotations are not made public. 
As a consequence, in LUNA16 teams could tune the parameters of their algorithm to show good performance on this particular data set, although the fact that LIDC-IDRI is a large set of scans from many different sources somewhat mitigates this risk. 
In order to allow performance comparison among different systems, we instructed participants that did not have their own training data to train their system in a particular cross-validation approach. Although this prevents some of the positive bias, positive bias may still remain if the design and architecture of a system are selected or optimized based on the full challenge data set. Previously published algorithms have also used LIDC-IDRI data to optimize the design of their systems. This may have influenced the design of algorithms used in LUNA16 as well.
Moreover, a cross-validation approach introduces some risks as people may make a mistake that goes unnoticed while carrying out a cross-validation experiment.
In fact, one team that originally participated in the challenge and reported excellent results had to withdraw because of a bug in the re-initialization of the network weights when starting training for the next fold in cross-validation. 
Unforeseen errors aside, allowing a cross-validation training procedure means that the presented systems are evaluated on test data while having been trained with data from the same sources (institutions, scanners, protocols). 
This may incur a positive bias in the reported results. This potential for bias is however also present in most, if not all, studies on the topic that have been previously published. Ruling out any possible bias is still an open problem for many machine learning competitions, even when the test data is not publicly available~(\cite{Blum15}).
On this note, it is important to further validate the performance and the generalizability of the systems using a completely independent validation data set.

Possible approaches for system improvement are as follows. Most of the non-nodular abnormalities, especially in lungs with many irregularities, were still detected. Although they may be clinically relevant, the algorithm should be able to differentiate these abnormalities. Classifying abnormalities into a subset of taxonomy, similar to \cite{Este17}, may be more useful for clinical usage. A more direct application would be to assign malignancy scores for all nodules detected by the system. This would further optimize the lung cancer screening workflow, for which less suspicious nodules would not have to undergo extensive screening protocol.

A future challenge could incorporate an even larger data set split into a training data set with annotations and a dedicated test data set for evaluation in which the reference standard is kept hidden. This still introduces a risk that teams can visually inspect the test data and the output of their system, notice false positives and false negatives and use that information to improve their performance. This could be circumvented by letting teams upload their algorithms, e.g.\ as machine executables or software containers, and evaluate these on test data that is not released publicly.

\section{Conclusions}

We have presented a web-based framework for a fair and automated evaluation of nodule detection algorithms, using the largest publicly available data set of chest CT scans in which nodules were annotated by multiple exert human readers. We have shown that combining classical candidate detection algorithms and analyzing these candidates with convolutional networks yields excellent results. 
Additionally, we have provided an update to the LIDC-IDRI reference standard which includes additional nodules found by CAD. The LUNA16 challenge will remain open for new submissions and can therefore be used as a benchmarking framework for future CT nodule CAD development.

\section*{Acknowledgements}

The authors acknowledge the National Cancer Institute and the Foundation for the National Institutes of Health and their critical role in the creation of the free publicly available LIDC-IDRI Database used in this study. This work was supported by a research grant from the Netherlands Organization for Scientific Research [project number 639.023.207].
A.~S\'o\~nora was supported by the Belgian Development Cooperation through the VLIR-UOS program with Universidad de Oriente in Santiago de Cuba. E.~Papavasileiou was supported by a PhD grant of the Research Foundation Flanders (FWO). 

\section*{References}
\footnotesize

\bibliographystyle{model2-names}
\bibliography{fullstrings,biblio}

\end{document}